\documentclass[journal]{IEEEtran}

\ifCLASSINFOpdf
\else
   \usepackage[dvips]{graphicx}
\fi
\usepackage{url}

\usepackage{graphicx}
\usepackage{algorithm}
\usepackage{algorithmic}
\usepackage{multirow}
\usepackage{pifont}
\usepackage{booktabs}
\usepackage{amssymb}
\usepackage{color}

\begin{document}

\title{Masked and Permuted Implicit Context Learning for Scene Text Recognition}

\author{Xiaomeng Yang, Zhi Qiao, Jin Wei, Dongbao Yang, Yu Zhou
% \IEEEmembership{Member, IEEE}
\thanks{This work was supported by the National Natural Science Foundation of China (Grant NO 62376266), and by the Key Research Program of Frontier Sciences, CAS (Grant NO ZDBS-LY-7024). \textit{(Corresponding author: Yu Zhou.)}}
\thanks{X. Yang, D. Yang and Y. Zhou are with the Institute of Information Engineering, Chinese Academy of Sciences, also with the School of Cyber Security, University of Chinese Academy of Sciences, Beijing 100089, China (e-mail: yangxiaomeng@iie.ac.cn; yangdongbao@iie.ac.cn; zhouyu@iie.ac.cn).}
\thanks{Z. Qiao is with the Tomorrow Advancing Life, Beijing 100080, China (e-mail: qiaozhi1@tal.com).}
\thanks{J. Wei is with the Lenovo Research, Beijing 100094, China (e-mail: weijin4@lenovo.com).}}

\markboth{Journal of \LaTeX\ Class Files, Vol. 00, No. 0, 2023}
{Shell \MakeLowercase{\textit{et al.}}: Bare Demo of IEEEtran.cls for IEEE Journals}
\maketitle

\begin{abstract}
Scene Text Recognition (STR) is difficult because of the variations in text styles, shapes, and backgrounds. Though the integration of linguistic information enhances models' performance, existing methods based on either permuted language modeling (PLM) or masked language modeling (MLM) have their pitfalls. PLM's autoregressive decoding lacks foresight into subsequent characters, while MLM overlooks inter-character dependencies. Addressing these problems, we propose a masked and permuted implicit context learning network for STR, which unifies PLM and MLM within a single decoder, inheriting the advantages of both approaches. We utilize the training procedure of PLM, and to integrate MLM, we incorporate word length information into the decoding process and replace the undetermined characters with mask tokens. Besides, perturbation training is employed to train a more robust model against potential length prediction errors. Our empirical evaluations demonstrate the performance of our model. It not only achieves superior performance on the common benchmarks but also achieves a substantial improvement of $9.1\%$ on the more challenging Union14M-Benchmark.
\end{abstract}


\IEEEpeerreviewmaketitle

\section{Introduction}

\IEEEPARstart{S}{cene} Text Recognition (STR) \cite{zhu2016scene, shi2018aster, baek2019wrong, qiao2020seed, long2021scene} aims to identify characters in natural scene images, which cannot only use visual information because of the variations of scene text in style, shape, and background. Therefore, integrating linguistic knowledge into STR models has attracted extensive attention. Prior works~\cite{yu2020towards, fang2021read, na2022multi} have achieved promising performance using external language models (LM). However, the independence between the vision model and LM may cause erroneous rectifications. Thus, an internal LM able to influence the visual space may be a better choice.

\begin{figure}[t]
\centering
\includegraphics[width=0.8\columnwidth]{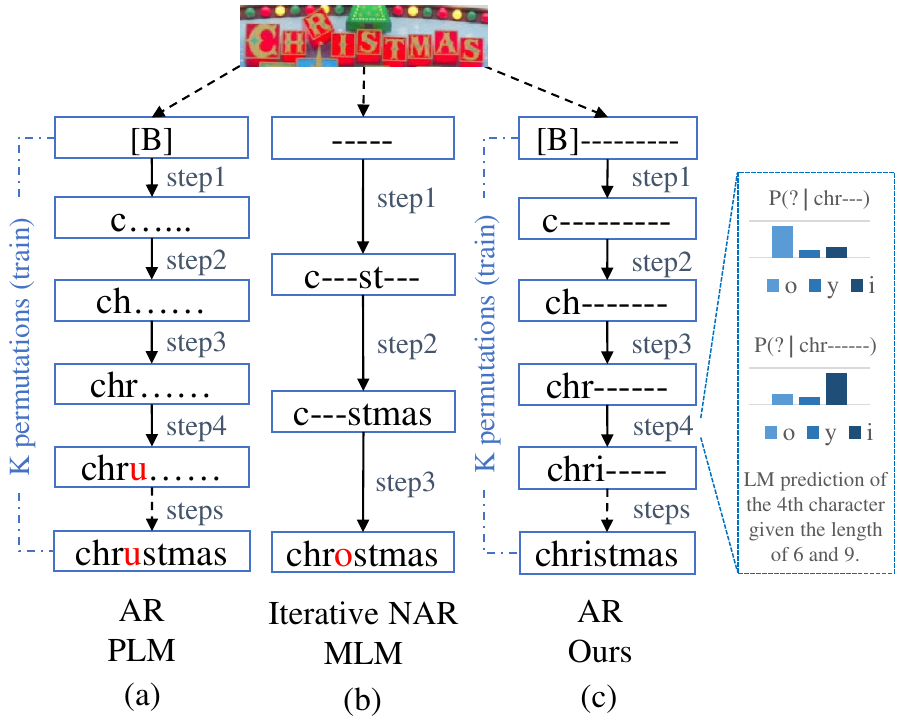} 
\caption{Decoding procedures of PLM, MLM, and our unified language modeling. ``-" in the figure represents the mask token used in decoding. The lack of global information and inadequate linguistic learning leads to misrecognition of PLM and MLM based methods.}
\label{decoding_procedure}
\end{figure}

The emergence of permuted language modeling (PLM) \cite{yang2019xlnet} and masked language models (MLM) \cite{devlin2018bert} have inspired the STR community to adopt novel decoding strategies.
PARSeq~\cite{bautista2022scene} uses an ensemble of permuted AR decodings to train the model but the serial procedure limits the LM's ability to anticipate future characters. As illustrated in Fig.~\ref{decoding_procedure}(a), when recognizing the fourth character in the word, the model lacks knowledge of the remaining characters, leading to the misrecognition. Conversely, MLM-based methods like PIMNet~\cite{qiao2021pimnet} offer global positional information about the entire text but may neglect dependency among each predicted character. 
As shown in Fig.~\ref{decoding_procedure}(b), this approach lacks contextual knowledge among characters predicted within the same iteration, such as ``m", ``a" and ``s" all determined in step 2. 
% Consequently, the internal LM learned by PIMNet is insufficient.
% and constrained by images present during training. 
In summary, both approaches have limitations. To improve STR performance, a more comprehensive context-learning method is needed. 
% Our goal is to enable the model to learn richer linguistic information implicitly during training, which would facilitate both AR and non-autoregressive (NAR) inference.

Based on the above analysis, we propose a masked and permuted implicit context learning network for STR (MPSTR). Inspired by \cite{song2020mpnet}, we combine PLM and MLM into a single decoder by rearranging tokens. 
Traditional iterative MLM decoders operate through cycles of prediction followed by re-masking. However, by defining a re-masking strategy in advance, the process can mirror AR decoding behaviors, but with additional context provided by the positional information of mask tokens. 
In our approach, the intrinsic permutations of PLM can serve as the ``re-masking strategy", determining the decoding order. 
Thus, the training procedure of our model follows PLM, but in each iteration, learnable mask tokens are appended behind the recognized characters to shape the context for the succeeding iteration. Additionally, since the single PLM decoder attends solely to previous tokens during inference, randomly predicted characters following the ending symbol cannot impact its prediction. However, incorporating MLM with redundant mask tokens demands length prediction. To solve this problem, we introduce an explicit length prediction using a simple learnable length token in the encoder. Specifically, the number of appended mask tokens is set to the number of un-transcribed characters in the current iteration, which enables more precise global information. For instance, the combination of the recognized ``chr" and appended six mask tokens shape the context for recognizing ``i" in Fig.~\ref{decoding_procedure}(c). Besides, to address potential inaccuracies in length prediction during inference, we employ a perturbation training strategy. Our approach intends to learn an advanced internal LM by uniquely incorporating mask tokens and length information to provide global linguistic knowledge. This differs from previous methods~\cite{xie2019aggregation,jiang2021reciprocal,li2022counting}, which directly use the length as an additional supervision.

We conduct extensive experiments to evaluate our method. Our MPSTR demonstrates superior performance on both six common benchmarks and the more challenging Union14M-Benchmark. The contributions are summarized as follows:
\begin{itemize}
    \item We propose a STR model that leverages the unification of PLM and MLM in a single decoder, inheriting their advantages and learning superior linguistic information.
    \item To integrate MLM, we employ a simple yet effective length prediction method using a learnable token in the encoder, along with perturbation training, which enhances the quality of global context information.
    \item Noticing mislabeled images in benchmarks, we conduct verification and provide the cleansed version in github.com/Xiaomeng-Yang/STR-benchmark-cleansed to ensure a more accurate evaluation of STR models.
    \item Our model achieves state-of-the-art performance on standard benchmarks and outperforms previous methods by a significant margin on the Union14M-Benchmark.
\end{itemize}

\begin{figure*}[t]
\centering
\includegraphics[width=\textwidth]{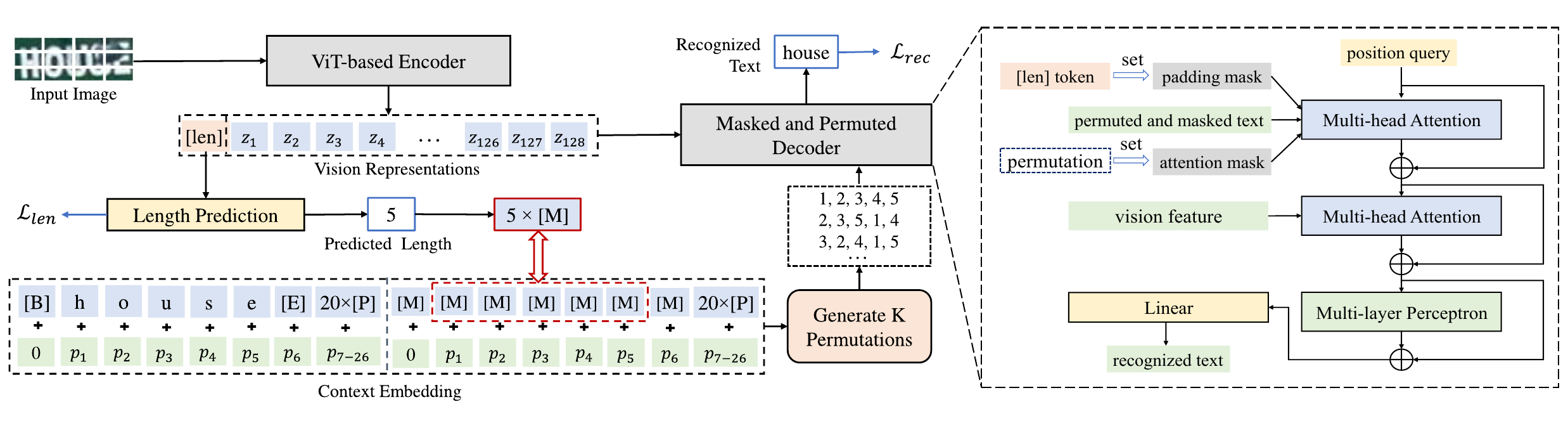} % Reduce the figure size so that it is slightly narrower than the column.
\caption{Architecture of the proposed method. $[B]$, $[E]$, $[P]$ and $[M]$ stands for the beginning-of-sequence, end-of-sequence, padding and mask tokens, respectively. The ViT-based encoder provides the text length using the $[len]$ token. Then, the predicted length number of mask tokens is appended. After K permutation operations, the masked and permuted text is input to the decoder for the corresponding prediction.}
\label{architecture}
\end{figure*}

\section{Method}
Our model contains a ViT-based encoder and a Masked and Permuted Decoder (MP-decoder) as shown in Fig.~\ref{architecture}. The encoder is used for visual feature extraction and length prediction.
% extracting visual features
% from the input image and predicting the length of the word. 
% Subsequently, 
The MP-decoder unifies MLM and PLM to decode the visual features into the final texts.
% predicted text 
% with contextual information. 
% This unified approach takes advantage of both MLM and PLM, allowing the model to effectively learn and leverage contextual information during the decoding process.

\subsection{ViT-based Encoder and Length Prediction}
The encoder is based on ViT but 
% As depicted in Figure 2, the input image $\mathbf{x}\in \mathbb{R}^{H\times W\times C}$ is divided into $P_w\times P_h$ patches. Then, $N = HW/(P_wP_h)$ patches are flattened into vectors and linearly projected into $D$-dimensional tokens. 
instead of the original $\left[class\right]$ token, we introduce a learnable $\left[len\right]$ token for length prediction.
% Additionally, position embeddings of equal dimension are added to each token to retain the positional information of the image patches. The final generated patch embedding vector is:
% \begin{equation}
% \mathbf{z}_{input} = \left[\mathbf{x}_{length};\mathbf{x}^1_p\mathbf{E};\mathbf{x}^2_p\mathbf{E};...;\mathbf{x}^N_p\mathbf{E}\right] + \mathbf{E}_{pos}
% \end{equation}
% where $\mathbf{x}_{length}\in \mathbb{R}^{1\times D}$ is the embedding of $\left[len\right]$ , $\mathbf{E}\in\mathbb{R}^{(P_hP_w)\times D}$is the patch embedding matrix and $\mathbf{E}_{pos}\in \mathbb{R}^{(N+1)\times D}$ is the position embedding.
Given $\mathbf{z}_{input}\in \mathbb{R}^{(N+1)\times D}$, the ViT encoder outputs the final embedding $\mathbf{z} = \left[\mathbf{z}_0,\mathbf{z}_1,\mathbf{z}_2,..., \mathbf{z}_N\right]$. We use $\mathbf{z}_0$ to predict the length of the word with two fully connected layers (FC) and an argmax layer. The word length prediction is treated as a $T$-class classification task, where $T$ represents the maximum possible length. 
% \begin{equation}
% L = {\rm argmax}({\rm FC}({\rm LN}(\mathbf{z}_0)))
% \end{equation}
Other learned tokens $\left[\mathbf{z}_1,\mathbf{z}_2,..., \mathbf{z}_N\right]$ are used as input vision representations for the MP-decoder.

\subsection{Masked and Permuted Language Modeling}
Let $\mathcal{Z}_T$ represent all possible permutations of a $T$-length sequence. A PLM-based method maximizes:
\begin{equation}
{\rm log}p(\mathbf{y}|\mathbf{x}) = \mathbb{E}_{z\sim\mathcal{Z}_T}\left[\sum^T_{t=1}{\rm log}p_\theta(y_{z_t}|\mathbf{y}_{\mathbf{z}_{<t}}, \mathbf{x})\right]
\end{equation}
where $z_t$ and $\mathbf{z}_{<t}$ denote the t-th character and the previous $t-1$ characters in the specific permutation $\mathbf{z}$. 
% In other words, a PLM-based decoder predicts the characters one by one in the order of each permutation.
% MLM-based methods can leverage the context derived from both mask tokens and unmasked characters.
For iterative MLM decoding, the model predicts all characters currently masked first and then re-masks some characters based on the predetermined strategy, such as easy-first~\cite{qiao2021pimnet}.
Let $\mathcal{K}_i$ represent masked positions in iteration $i$, $I$ be the number of iterations, $\mathbf{y}_{\setminus\mathcal{K}_i}$ be determined characters and $\mathbf{M}_{\mathcal{K}_i}$ be mask tokens. An iterative MLM-based NAR decoder maximizes:
\begin{equation}
{\rm log}p(\mathbf{y}|\mathbf{x}) \approx \sum^I_{i = 1}\sum_{k\in\mathcal{K}_i}{\rm log}p_\theta(y_k|\mathbf{y}_{\setminus\mathcal{K}_i}, \mathbf{M}_{\mathcal{K}_i}, \mathbf{x})
\end{equation}

% \begin{figure}[]
% \centering
% \includegraphics[width=0.9\columnwidth]{images/decoder.pdf} % Reduce the figure size so that it is slightly narrower than the column.
% \caption{The architecture of Masked-Permuted Decoder.}
% \label{decoder}
% \end{figure}

To unify PLM and MLM, we analyze the $c$-th step of AR decoding in a chosen permutation of PLM, where the characters $\mathbf{y}_{\mathbf{z}<c}$ are already generated. In this situation, the MLM decoder masks the characters $\mathbf{y}_{\mathbf{z}\ge c}$ and will predict them depending on previously deduced tokens $\mathbf{y}_{\mathbf{z}<c}$ and the subsequent mask tokens $\mathbf{M}_{\mathbf{z}\ge c}$.
% It is crucial to note that while the MLM simultaneously decodes all masked characters in parallel, the contextual insights harvested by each masked token resemble the scenario of decoding them sequentially in a specific iteration. This is not to imply that subsequent tokens rely on preceding ones. Instead, it suggests that what could be predicted in a singular step is spread across multiple steps. Consider Figure~\ref{decoding_procedure}(b) as an illustrative example. During the second iteration, one can sequentially derive the prediction for ``m", ``a" and ``s" using the same context ``c- - -st- - -". The contextual information leveraged for these three tokens is analogous to predicting them concurrently. 
% Unlike the conventional iterative MLM decoder, which iteratively predicts and then re-masks tokens, an alternative approach emerges, directly predicting specific masked tokens in a pre-established masking order. If we equate the MLM iteration to the maximum word length, this process mirrors AR decoding. However, the distinguishing factor is the global context drawn from the positional attributes of the masked tokens. In this paradigm, the selected permutation represents the re-masking strategy. 
Thus, we can unify PLM and MLM into one decoder such that when generating the next character autoregressively in each permutation, the unified MP-decoder should depend on both the predicted part and several mask tokens whose number is equal to the number of remaining undetermined characters. The unified training objective of the MP-decoder is formulated as:
\begin{equation}\label{eq:mp}
\mathbb{E}_{z\sim\mathcal{Z}_T}\left[\sum_{t=c}^T{\rm log}p_\theta(y_{z_t}|\mathbf{y}_{\mathbf{z}_{<t}},\mathbf{M}_{\mathbf{z}_{\ge c}}, \mathbf{x})\right],
\end{equation}
% which inherits the ensemble of permuted AR decoding training process from PLM and mask tokens' information for unpredicted characters like MLM.

\subsection{Masked and Permuted Decoder}

\begin{table}[]
    \centering
    \caption{The attention mask $\mathbf{m}_{attn}$ for permutation$[1,3,2]$. $1$ means the token in the table header is masked when decoding.}
    \label{tab:attn_mask}
    \begin{tabular}{c|ccccc|ccccc}
         & $\underline{B}$ & $y_1$ & $y_2$ & $y_3$ & $\underline{E}$ & $\underline{M}$ & $\underline{M}$ & $\underline{M}$ & $\underline{M}$ & $\underline{M}$ \\
        \hline
        $y_1$ & 0 & 1 & 1 & 1 & 1 & 1 & 0 & 0 & 0 & 0 \\
        $y_2$ & 0 & 0 & 1 & 0 & 1 & 1 & 1 & 0 & 1 & 0\\
        $y_3$ & 0 & 0 & 1 & 1 & 1 & 1 & 1 & 0 & 0 & 0\\
        $\underline{E}$ & 0 & 0 & 0 & 0 & 1 & 1 & 1 & 1 & 1 & 0\\
       \hline
    \end{tabular}
\end{table}

The MP-decoder adopts two Multi-Head Cross-Attention (MHCA) modules. The first MHCA receives position query $\mathbf{p}$ with length $T+1$ for an additional $[EOS]$ token:
\begin{equation}
\mathbf{h}_c =\mathbf{p}+{\rm MHCA(\mathbf{p}, \mathbf{c}, \mathbf{c}, \mathbf{m}_{attn}, \mathbf{m}_{pad})}
\end{equation}
where $\mathbf{c} = \left[\mathbf{c}_{word};\mathbf{c}_{mask}\right] \in \mathbb{R}^{2(T+2)\times D}$ is the concatenation of word context embeddings $\mathbf{c}_{word} \in \mathbb{R}^{(T+2)\times D}$ and mask context embeddings $\mathbf{c}_{mask} \in \mathbb{R}^{(T+2)\times D}$, both of which are added with positional information as shown in Fig.~\ref{architecture}. $\mathbf{m}_{attn}\in\mathbb{R}^{(T+1)\times 2(T+2)}$ and $\mathbf{m}_{pad}\in\mathbb{R}^{2(T+2)}$ are the attention mask and padding mask, respectively. Since we cannot train on all $L!$ permutations, we choose $K$ permutations from $\mathcal{Z}_L$, and the chosen permutations are informed to the decoder through the attention mask.
The attention mask $\mathbf{m}_{attn}=\left[\mathbf{m}_{wattn};\mathbf{m}_{mattn}\right]$ is generated uniquely for each chosen permutation according to equation (\ref{eq:mp}), where $\mathbf{m}_{wattn}$ and $\mathbf{m}_{mattn}$ are attention masks for $\mathbf{c}_{word}$ and $\mathbf{c}_{mask}$, respectively. At a certain step of AR decoding, the unpredicted part of the word context embeddings and the mask context embeddings corresponding to the predicted tokens are masked, e.g. the example in Table~\ref{tab:attn_mask}. When decoding the character $y_3$, $[B]$ and $y_1$ in $c_{word}$, and the last three $[M]$ in $c_{mask}$ could be attended. The padding mask is determined by the predicted word length $L$ with the first $L+2$ tokens unmasked for both $\mathbf{c}_{word}$ and $\mathbf{c}_{mask}$. In other words, $\mathbf{m}_{pad} = \left[\mathbf{m}_{wpad};\mathbf{m}_{mpad}\right]$, where $\mathbf{m}_{wpad}=\mathbf{m}_{mpad}$. For the word ``house" with length 5, $\mathbf{m}_{wpad}=\mathbf{m}_{mpad}=\left[1,1,1,1,1,1,1,0,...,0\right]\in\mathbb{R}^{T+2}$. 
The second MHCA computes the attention between $\mathbf{h}_c$ and image visual features $\mathbf{z}_{v}$ obtained from the ViT encoder. After that, the prediction is ultimately achieved by processing the output logits through an MLP and an FC layer.

\begin{table*}[t]
    \centering
    \caption{Recognition results on Union14M-Benchmark. ``N" and ``A" represent NAR and AR decodings, respectively.}
    \label{union}
    \begin{tabular}{l|ccccccc|c|c}
    \hline
    Method & Curve & Multi-Oriented & Artistic & Contextless & Salient & Multi-Words & General & Avg & Incomplete ($\downarrow$) \\
    \hline
    CRNN~\cite{shi2016end} & 19.4 & 4.5 & 34.2 & 44.0 & 16.7 & 35.7 & 60.4 & 30.7 & \textbf{0.9}\\
    SAR~\cite{li2019show} & 68.9 & 56.9 & 60.6 & 73.3 & 60.1 & 74.6 & 76.0 & 67.2 & 2.1\\
    SATRN~\cite{wang2020decoupled} & 74.8 & 64.7 & 67.1 & 76.1 & 72.2 & 74.1 & 75.8 & 72.1 & \textbf{0.9}\\
    SRN~\cite{yu2020towards} & 49.7 & 20.0 & 50.7 & 61.0 & 43.9 & 51.5 & 62.7 & 48.5 & 2.2\\
    ABINet~\cite{fang2021read} & 75.0 & 61.5 & 65.3 & 71.1 & 72.9 & 59.1 & 79.4 & 69.2 & 2.6\\
    VisionLAN~\cite{wang2021two} & 70.7 & 57.2 & 56.7 & 63.8 & 67.6 & 47.3 & 74.2 & 62.5 & \underline{1.3}\\
    SVTR~\cite{du2022svtr} & 72.4 & 68.2 & 54.1 & 68.0 & 71.4 & 67.7 & 77.0 & 68.4 & 2.0\\
    MATRN~\cite{na2022multi} & 80.5 & 64.7 & 71.1 & 74.8 & 79.4 & 67.6 & 77.9 & 74.6 & 1.7\\
    MAERec-S~\cite{jiang2023revisiting} & 75.4 & 66.5 & 66.0 & 76.1 & 72.6 & 77.0 & 80.8 & 73.5 & 3.5\\
    MAERec-B~\cite{jiang2023revisiting} & 76.5 & 67.5 & 65.7 & 75.5 & 74.6 & \underline{77.7} & 81.8 & 74.2 & 3.2\\
    \hline
    % MAERec-S & 81.4 & 71.4 & 72.0 & 82.0 & 78.5 & 82.4 & 82.5 & 78.6 & 2.7\\
    % MAERec-B~\cite{} & 88.8 & 83.9 & 80.0 & 85.5 & 84.9 & 87.5 &  85.8 & 85.2 & 2.6\\
    % \hline
    MPSTR$_{N}$ (Ours) & 82.4 & 80.9 & 70.6 & 77.0 & 79.1 & 71.5 & 81.4 & 77.6 & 1.4\\
    MPSTR$_{A}$ (Ours) & 84.0 & 83.1 & 71.8 & 78.3 & 80.1 & 76.7 & 82.4 & 79.5 & \underline{1.3}\\
    MPSTR-B$_{N}$ (Ours) & \underline{85.9} & \underline{85.0} & \underline{74.6} & \underline{80.4} & \underline{81.7} & 75.2 & \underline{83.2} & \underline{80.9} & 1.5\\
    MPSTR-B$_{A}$ (Ours) & \textbf{87.8} & \textbf{87.4} & \textbf{77.2} & \textbf{83.1} & \textbf{83.6} & \textbf{82.2} & \textbf{84.3} & \textbf{83.7} & 1.5\\
    \hline
    \end{tabular}
\end{table*}
 
% \begin{equation}
% \mathbf{h}_i =\mathbf{h}_c+{\rm MHCA}(\mathbf{h}_c, \mathbf{z}_v, \mathbf{z}_v)
% \end{equation}
% \begin{equation}
% \mathbf{y} ={\rm FC}(\mathbf{h}_i+{\rm MLP}(\mathbf{h}_i))
% \end{equation}

\subsection{Training Objective and Strategies}
The objective function of our proposed method is:
\begin{equation}
\mathcal{L} = \lambda \mathcal{L}_{len} + (1-\lambda) \mathcal{L}_{rec}
\end{equation}
where $\mathcal{L}_{len}$ represents the loss of length prediction, and $\mathcal{L}_{rec}$ stands for the recognition loss. The parameter $\lambda$ is employed to balance the two losses. We use the cross-entropy loss for $\mathcal{L}_{len}$. 
% Let $p_{len}$ and $g_{len}$ denote the predicted word length and ground truth word length respectively, and $L_{ce}$ represent the cross-entropy loss.
% \begin{equation}
% \mathcal{L}_{len} = L_{ce}(p_{len}, g_{len})
% \end{equation}
For recognition loss, we compute the mean of the cross-entropy losses
% between predicted text $y_k$ and ground truth label $\hat{y}$ 
for $K$ chosen permuted decoding orders.
During training, we employ a teacher-forcing strategy, using the ground truth length to set the masks.
% , which aids in learning advanced contextual information. 
For inference, we use the predicted length. Additionally, to enhance the robustness of the decoder against the potential length prediction errors during inference, we adopt a perturbation training strategy.
For each batch, we perturb the ground truth length for a randomly selected subset of images, either by adding or subtracting one. 
% For the selected images, the perturbed length is utilized to determine the number of mask tokens
% while other images still using the gourd truth length.

\section{Experiments}
\subsection{Datasets}
We use three groups for training: (1) Synthetic data (S): MJSynth~\cite{jaderberg2016reading} and SynthText~\cite{gupta2016synthetic}, (2) Real images~\cite{bautista2022scene} (R), and (3) Union14M-L's training set~\cite{jiang2023revisiting}. For evaluation, we assess the models on six benchmarks: IIIT5k-word (IIIT5k)~\cite{mishra2012scene}, ICDAR 2013 (IC13)~\cite{karatzas2013icdar}, Street View Text (SVT)~\cite{wang2011end}, ICDAR 2015 (IC15)~\cite{karatzas2015icdar}, SVT-Perspective (SVTP)~\cite{phan2013recognizing}, and CUTE80 (CUTE)~\cite{risnumawan2014robust}. However, there are some mislabeled images, so we perform further verification based on \cite{baek2019wrong}. Additional details can be found in our supplementary file. We also test our models on the Union14M-Benchmark~\cite{jiang2023revisiting}.

\subsection{Implementation Details}

\begin{table}[t]
    \setlength\tabcolsep{2.5pt}
    \scriptsize
    \centering
    \caption{Recognition results on cleansed six benchmarks. * indicates that the reproduced models. MGP-STR$^\dagger$ represents the small configuration. S$^1$ LM pretrained on WikiText-103~\cite{merity2017pointer}.}
    \label{sota_comparison}
    \begin{tabular}{l|c|ccc|ccc|c|c}
        \hline
        \multirow{2}{*}{Method} & Train & IIIT5K & IC13 & SVT & IC15 & SVTP & CUTE & Avg & Time\\
        & Data & 3000 & 857 & 647 & 1811 & 645 & 288 & 7248 & (ms/img)\\
        \hline
        TRBA~\cite{baek2019wrong} & S & 90.5 & 93.3 & 87.3 & 79.0 & 78.5 & 72.9 & 85.9 & 16.7\\
        ViTSTR~\cite{atienza2021vision} & S & 88.9 & 92.6 & 87.2 & 80.5 & 82.3 & 80.9 & 86.2 & 7.92\\ 
        SRN~\cite{yu2020towards} & S & 92.2 & 95.2 & 89.6 & 81.7 & 85.1 & 88.5 & 88.9 & 12.1\\
        VisionLAN~\cite{wang2021two}& S & 96.5 & 96.3 & 90.4 & 85.4 & 85.7 & 88.9 & 91.9 & 14.8\\
        PIMNet*~\cite{qiao2021pimnet} & S & 95.5 & 95.9 & 92.0 & 86.5 & 87.9 & 90.6 & 92.1 & 15.6\\
        ABINet~\cite{fang2021read} & S$^1$ & 97.0 & \textbf{97.0} & 93.4 & 87.3 & \textbf{89.6} & 89.2 & 93.3 & 23.4\\
        MGP-STR$^\dagger$~\cite{wang2022multi} & S & 95.6 & 96.6 & 93.0 & 87.8 & 89.0 & 88.5 & 92.7 & 9.37\\
        PARSeq$_N$*~\cite{bautista2022scene} & S & 96.0 & 96.6 & 92.4 & 86.8 & 88.7 & 90.6 & 92.6 & 11.7\\
        PARSeq$_A$*~\cite{bautista2022scene} & S & 96.6 & \textbf{97.0} & 92.7 & 87.8 & \underline{89.3} & \textbf{91.7} & 93.3 & 17.5\\
        LevOCR~\cite{da2022levenshtein} & S & \underline{97.1} & 96.7 & \textbf{94.4} & \underline{88.4} & \textbf{89.6} & \underline{90.6} & \underline{93.7} & 60.5 \\
        \hline
        MPSTR$_N$ (Ours) & S & 96.8 & 96.4 & 92.6 & 88.2 & 88.5 & \textbf{91.7} & 93.3 & 12.6\\
        MPSTR$_A$ (Ours) & S & \textbf{97.2} & \underline{96.9}  & \underline{93.2} & \textbf{89.0} & \textbf{89.6} & \textbf{91.7} & \textbf{93.9} & 19.1\\
        \hline
        PIMNet*~\cite{qiao2021pimnet} & R & 97.7 & 97.1 & 96.5 & 90.9 & 92.9 & 95.1 & 95.3 & 15.6\\
        PARSeq$_N$~\cite{bautista2022scene} & R & 98.1 & 97.9 &  97.2 & 92.8 & 94.0 & 98.6 & 96.3 & 11.7 \\
        PARSeq$_A$~\cite{bautista2022scene} & R & 98.9 & \underline{98.3} & 97.5 & \underline{93.7} & \underline{95.7} & \underline{98.6} & 97.1 & 17.5\\
        MPSTR$_N$ (Ours) & R & \underline{99.0} & \textbf{98.4} & \underline{98.3} & \underline{93.7} & 95.5 & \underline{98.6} & \underline{97.2} & 12.6\\
        MPSTR$_A$ (Ours) & R & \textbf{99.2} & \underline{98.3} & \textbf{98.5} & \textbf{93.9} & \textbf{96.1} & \textbf{99.0} & \textbf{97.4} & 19.1\\
        \hline
    \end{tabular}
\end{table}

We employ the small and base DeiT~\cite{touvron2021training} for MPSTR and MPSTR-B with $128\times32$ images and $8\times4$ patches. Our model uses 36-char for common benchmarks and 91-char for Union14M. Data augmentations are consistent with PARSeq~\cite{bautista2022scene}. We employ the Adam optimizer with the 1cycle~\cite{smith2019super} scheduler for the initial $85\%$ iterations and the Stochastic Weight Averaging scheduler for others. Parameters are set to $T=25$, $K=12$ and $\lambda=0.25$. The models are trained for $254,520$ iterations on S and R, and 10 epochs on Union14M-L on NVIDIA Tesla V100. One Refinement is used employing the cloze mask for AR and two for NAR like \cite{bautista2022scene}. Word accuracy is assessed ignoring cases and symbols.

\subsection{Comparison with State-of-the-Art}
\subsubsection{Union14M-Benchmark}
% Since STR performance is saturated on common benchmarks, 
We highlight our method's performance on the Union14M-Benchmark. 
As detailed in Table~\ref{union}, our models surpass the previous best approach by an average of $9.1\%$. Notably, on the multi-oriented dataset, our model exceeds the previous best by $19.2\%$, attributable to our robust linguistic modeling. Besides, while our approach lags behind language-free methods on incomplete images due to semantic dependencies, the word length prediction mitigates this impact, so it is better than other language-based methods.

\subsubsection{Common Benchmarks}
We evaluate models on cleansed benchmarks with reproduction for PARSeq~\cite{bautista2022scene} using its official configuration and PIMNet~\cite{qiao2021pimnet} using the DeiT-small encoder. As shown in Table~\ref{sota_comparison}, our approach achieves comparable results when trained on both S and R. It surpasses LevOCR~\cite{da2022levenshtein} by $0.2\%$, but with only a third of inference time. Compared with the PLM-based PARSeq, our AR model improves accuracy over PARSeq$_A$ by $0.6\%$ and $0.3\%$, and our NAR model beats PARSeq$_N$ with boosts of $0.7\%$ and $0.9\%$. MPSTR$_N$ even matches the performance of PARSeq$_A$ with a latency reduction of about 5ms. In comparison with MLM-based methods, MPSTR$_N$ betters PIMNet by $1.0\%$ and $1.5\%$. 
% These results demonstrate the effectiveness and efficiency of our approach. 

\subsection{Ablation Study}
\subsubsection{Language Modeling Method}
\begin{table}[t]
    \centering
    \caption{Ablation study on language modeling strategy. Baseline is an AR model. * model evaluated with the GT lengths. (R/S) indicates the training data.}
    \label{tab:LM}
    \begin{tabular}{l|cc|cc}
        \hline
        Method & AR (R) & NAR (R) & AR (S) & NAR (S) \\
        \hline
        baseline & 96.63 & 0.00 & 93.12 & 0.00\\
        PLM & 97.10 & 96.33 & 93.25 & 92.58\\
        PLM+MLM & \textbf{97.43} & \textbf{97.21} & \textbf{93.86} & \textbf{93.29} \\
        \hline
        PLM+MLM* & 97.71 & 97.67 & 94.44 & 94.18 \\
       \hline
    \end{tabular}
\end{table}
As shown in Table~\ref{tab:LM}, the baseline left-to-right AR model makes NAR inference inapplicable. Incorporating PLM facilitates NAR decoding by learning linguistic relationships. MLM further enhances both AR and NAR accuracy thanks to its global contextual insights. 
% Notably, the NAR models benefit more significantly from the global positioning details offered by MLM. When used by itself, PLM does not provide much meaningful information beyond the initial token, leading to potential errors like the redundant ``t" predicted by PARSeq$_N$ in Figure~\ref{visualization}.
An upper bound is given in the final row. Our integrated language modeling has superior context comprehension, driven by global context and comprehensive character dependencies.

\subsubsection{MLM Integration}
\begin{table}[t]
    \setlength\tabcolsep{3pt}
    \centering
    \caption{Ablation study of MLM integration.}
    \label{tab:length_usage}
    \begin{tabular}{c|c|cc|cc}
        \hline
        Mask Token & Length& AR (R) & NAR (R) & AR (S) & NAR (S) \\
        \hline
        \ding{55} & \ding{55} & 97.10 & 96.33 & 93.25 & 92.58\\
        \ding{51} & \ding{55} & 96.83 & 53.42 & 93.36 & 36.58 \\
        \ding{55} & \ding{51} & 96.67 & 96.39 & 93.23 & 92.54 \\
        \ding{51} & \ding{51} & \textbf{97.43} & \textbf{97.21} & \textbf{93.86} & \textbf{93.29}  \\
       \hline
    \end{tabular}
\end{table}

\begin{table}[t]
    \centering
    \caption{Ablation study about length prediction methods.}
    \label{tab:length_prediction}
    \begin{tabular}{l|cc|cc||cc}
        \hline
        \multirow{2}{*}{Method} & \multicolumn{2}{|c|}{Real Data} & \multicolumn{2}{|c||}{Synthetic Data} & \multicolumn{2}{c}{Length }\\
         & AR & NAR & AR & NAR & Real & Synth \\
        \hline
        Char & 96.74 & 96.04 & 92.80 & 91.63 & 98.44 & 95.41\\
        Word & \textbf{97.43} & \textbf{97.21} & \textbf{93.86} & \textbf{93.29} & \textbf{98.79}& \textbf{96.41} \\
       \hline
    \end{tabular}
\end{table}

\begin{table}[!h]
    \centering
    \caption{MPSTR$_N$'s reconigtion results vs perturbation ratio.}
    \label{tab:perutrbation}
    \begin{tabular}{c|cccccc}
        \hline
        Data & 0\% & 25\% & 33\% & 50\% & 66\% & 75\% \\
        \hline
        S & 92.09 & 92.47 & 92.55 & 92.74 & \textbf{93.10} & 92.80\\
        R & 96.72 & 96.81 & \textbf{97.21} & 96.73 & - & -\\
       \hline
    \end{tabular}
\end{table}

Table~\ref{tab:length_usage} reveals that merely introducing mask tokens without length supervision prohibits NAR inference. This is expected, as our unified likelihood defined in equation (\ref{eq:mp}) necessitates the number of unpredicted tokens. Solely adding the predicted length embedding to the first MHCA's query also cannot provide satisfying results because of imprecise length predictions. It suggests that both mask tokens and length are required to incorporate MLM.

\subsubsection{Word Length Prediction}
Two methods of length prediction are available: character-level prediction, which determines the presence of each character, and word-level prediction, which directly estimates the word's length. We use a parallel decoder for binary classification at each position for character-level prediction, and the length token for word-level prediction. As shown in Table~\ref{tab:length_prediction}, the word-level approach performs better since it utilizes global information.
% Therefore, we utilize the word-level $[len]$ token prediction strategy.

\subsubsection{Perturbation training}
% To examine the impact of perturbation training
% % on the decoder's robustness for length prediction
% , we performed an ablation study on the percentage of images trained with perturbed length. 
Table~\ref{tab:perutrbation} reveals that models trained on S need a higher proportion of perturbed images ($66\%$) due to their regularity, while models trained on R perform best with just $33\%$ perturbed images.

% \subsection{Qualitative Analysis}
% \begin{figure}[t]
% \centering
% \includegraphics[width=1\columnwidth]{images/attention_mapnew.pdf} % Reduce the figure size so that it is slightly narrower than the column.
% % \caption{Visualization of the first MHCA's attention weights of MPSTR$_A$.}
% % \label{attention_map}
% % \end{figure}

% We visualize the MPSTR$_A$ attention weights of the permuted and masked text in MP-decoder for two challenging examples that PARseq fails in Figure~\ref{attention_map}. As shown in this figure, MP-decoder could obtain not only previous characters' information, but also the information provided by padding mask tokens with specific length. 
% Through training on several permutations of the original text, MPSTR could learn better internal language model with global linguistic information and build dependency among every characters. Figure~\ref{visualization} shows that MPSTR provides more robust results on challenging examples compared with PARSeq.

% \begin{figure}[t]
% \centering
% \includegraphics[width=1\columnwidth]{images/visualization_1.pdf} % Reduce the figure size so that it is slightly narrower than the column.
% \caption{Challenging examples that PARSeq fails (upper) but our MPSTR (bottom) performs better. ``mari-iotttttt..." stands for the redundancy of the last character "t".}
% \label{visualization}
% \end{figure}

\section{Conclusion}
We propose MPSTR to overcome the limitations of PLM and MLM, learning a more effective internal LM with advanced linguistic information. Additionally, word-level text length prediction with perturbation training is introduced to effectively integrate the MLM method. Our method achieves SOTA performance on most public benchmarks and shows consistent improvements on the challenging Union14M-Benchmark, validating its effectiveness. 

% Furthermore, the  demonstrate the generalization capabilities of our proposed method.

\bibliographystyle{IEEEtran}
\bibliography{mpstr}

\clearpage
\section*{Appendix}
\subsection{Model Evaluation}
We utilize AR decoding with one refinement iteration and NAR decoding with two refinement iterations. The refinement is performed using the cloze mask. 

The NAR decoding generates all tokens simultaneously with all position queries $\mathbf{p} = [\mathbf{p}_1, \mathbf{p}_2, ..., \mathbf{p}_{T+1}]$, where $T$ is the maximum length, and $\mathbf{p}_T+1$ is added for the end-of sequence (eos) token. For NAR inference, we provide the beginning token $[B]$ along with $L+2$ mask tokens $[M]$, where $[M]_0$ and $[M]_{L+1}$ represent the corresponding mask tokens for beginning token and eos token, as context.

The AR decoding generates one new token per iteration from left to right. The attention mask for the AR decoding procedure is shown in Table~\ref{tab:ar_mask}. For the first iteration, it predicts the first token $y_1$ with the position query $\mathbf{p}_1$, using the context set to $[B]$ concatenated with $L+1$ mask tokens. For the succeeding iteration $i$, the context is set to the determined tokens (i.e. $[B]$ together with the determined characters $y_1$ to $y_{i-1}$) concatenated with the mask tokens for the unpredicted part (i.e. $[M]_i$ to $[M]_{L+1}$). We use $\mathbf{p}_i$ as the position query. 

Before each refinement, the length is re-determined based on the predicted text from the previous iteration. When decoding a character $y_i$, the query $\mathbf{p}_i$ can attend to all information from the previous iteration except itself. Therefore, determined tokens and one mask token for itself are set as the context, as illustrated in Table~\ref{tab:cloze_mask}. 
% For examples, when refining $y_2$, the attention mask of $\left[[B], y_1, y_2, ..., y_L, [E], [M]_0, [M]_1, [M]_2, ..., [M]_L, [M]_{L+1}\right]$ is $[0, 0, 1, 0, 0, 0, 1, 1, 0, 1, 1, 1]$.
All tokens are refined simultaneously.

\begin{table}[t]
    \setlength\tabcolsep{2.5pt}
    \setcounter{table}{7}
    \centering
    \caption{The AR decoding attention mask. 1 Means masked.}
    \label{tab:ar_mask}
    \begin{tabular}{c|cccccc|cccccc}
        \hline
         & $[B]$ & $y_1$ & $y_2$ & ... & $y_L$ & $[E]$ & $[M]_0$ & $[M]_1$ & $[M]_2$ & ... & $[M]_L$ &$[M]_{L+1}$ \\
        \hline
        $y_1$ & 0 & 1 & 1 & 1 & 1 & 1 & 1 & 0 & 0 & 0 & 0 & 0 \\
        $y_2$ & 0 & 0 & 1 & 1 & 1 & 1 & 1 & 1 & 0 & 0 & 0 & 0\\
        $...$ & 0 & 0 & 0 & 1 & 1 & 1 & 1 & 1 & 1 & 0 & 0 & 0\\
        $y_L$ & 0 & 0 & 0 & 0 & 1 & 1 & 1 & 1 & 1 & 1 & 0 & 0\\
        $[E]$ & 0 & 0 & 0 & 0 & 0 & 1 & 1 & 1 & 1 & 1 & 1 & 0\\
       \hline
    \end{tabular}
\end{table}

\begin{table}[t]
    \setlength\tabcolsep{2.5pt}
    \centering
    \caption{The cloze attention mask for refinement.}
    \label{tab:cloze_mask}
    \begin{tabular}{c|cccccc|cccccc}
        \hline
         & $[B]$ & $y_1$ & $y_2$ & ... & $y_L$ & $[E]$ & $[M]_0$ & $[M]_1$ & $[M]_2$ & ... & $[M]_L$ & $[M]_{L+1}$ \\
        \hline
        $y_1$ & 0 & 1 & 0 & 0 & 0 & 0& 1 & 0 & 1 & 1 & 1 & 1 \\
        $y_2$ & 0 & 0 & 1 & 0 & 0 & 0& 1 & 1 & 0 & 1 & 1 & 1 \\
        $...$ & 0 & 0 & 0 & 1 & 0 & 0& 1 & 1 & 1 & 0 & 1 & 1 \\
        $y_L$ & 0 & 0 & 0 & 0 & 1 & 0& 1 & 1 & 1 & 1 & 0 & 1 \\
        $[E]$ & 0 & 0 & 0 & 0 & 0 & 1& 1 & 1 & 1 & 1 & 1 & 0 \\
       \hline
    \end{tabular}
\end{table}

\subsection{Benchmark Cleansing Details}

\begin{figure}[t]
\setcounter{figure}{2}
\centering
\includegraphics[width=0.8\columnwidth]{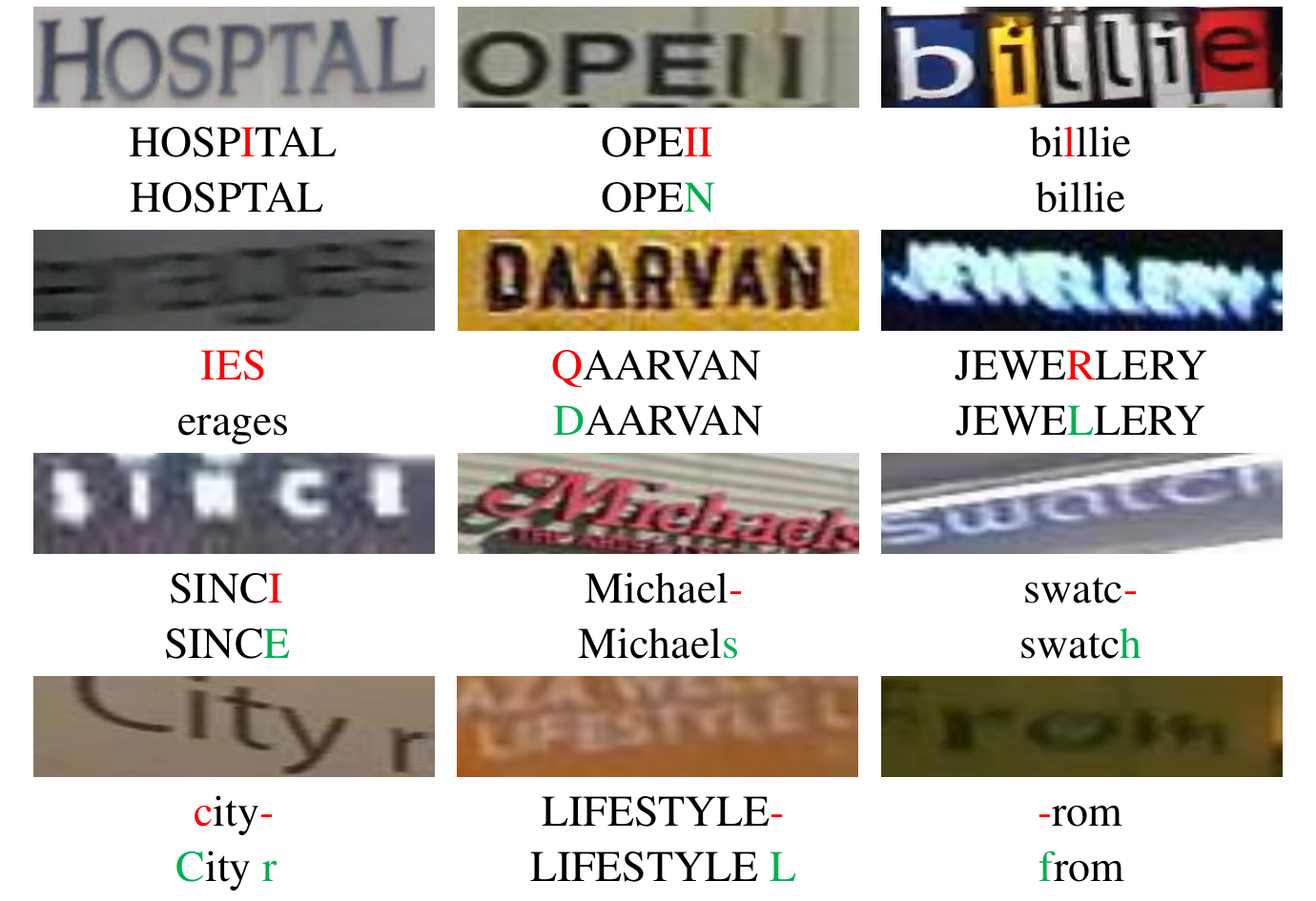} % Reduce the figure size so that it is slightly narrower than the column.
\caption{Examples of label noise in benchmark datasets are presented. Characters that are mislabeled or missing are highlighted in red (top), while the corrected labels are provided below.}
\label{label noise}
\end{figure}

Despite the widespread use of these existing benchmarks in scene text recognition, we have noticed that there are mislabeled images in the benchmark, as illustrated by the examples in Fig.~\ref{label noise}. State-of-the-art STR models have achieved an accuracy over $95\%$. However, the common benchmarks still contain noise, and future work might continue to evaluate models on these noisy benchmarks. We consider this a serious issue, and it is why we embarked on the task of cleaning the benchmark datasets. We conduct further verification of these labels. For example, when examining the image ``OPEII", we identified the final two ``II" characters as a single entity ``N" based on the character pitch. Even if we were to consider the last two characters as two ``I"s, the shape does not match that of a capital ``I". And for ``SINCI", the shape of the last character is not like the second clear ``I". The cleansed benchmark datasets are provided to ensure a more accurate evaluation of scene text recognition models.

\subsection{More Ablation Study Analysis}

\subsubsection{Permutation Number}

\begin{table}[t]
    \caption{Recognition accuracy of MPSTR trained using real datasets on the benchmark vs the number of permutations used ($K$) for training. No perturbation is applied for these models. AR acc. and NAR acc. are evaluated without refinement. For cloze acc., we use GT label as the initial prediction.}
    \label{tab:permutation}
    \centering
    \begin{tabular}{r|ccc}
        \hline
        K & AR acc. & NAR acc. & cloze acc. \\
        \hline
       1 & 96.18 & 0.00 & 63.23 \\
       2 & 96.58 & 1.32 & 97.54 \\
       6 & 96.23 & 96.10 & 98.00\\
       12 & \textbf{96.65} & 96.32 & \textbf{98.08}\\
       18 & 96.45 & \textbf{96.34} & 97.97 \\
       \hline
    \end{tabular}
\end{table}

Results presented in Table~\ref{tab:permutation} highlight that the NAR decoding procedure encounters difficulties with a limited number of permutations ($K=1,2$). This is because there is insufficient contextual information. In contrast, AR decoding shows robust performance, primarily due to the alignment between training and inference, augmented by the length supervision. As $K$ grows, there is a noticeable improvement in both AR and NAR performances up to $K\le12$. It is worth noting that our MPSTR demands more permutations for optimal performance than the PLM-based PARseq. This is attributed to the enriched global linguistic information derived from the padding of mask tokens.

\subsubsection{Balance Weight $\lambda$}
\begin{table}[!t]
    \caption{MPSTR$_A$'s recognition results on the benchmark vs loss balance hyper-parameter $\lambda$ without perturbation training.}
    \label{tab:perutrbation}
    \centering
    \begin{tabular}{c|ccccc}
        \hline
        $\lambda$ & 0.15 & 0.20 & 0.25 & 0.33 & 0.50 \\
        \hline
        Acc & 96.53 & 96.60 & \textbf{96.72} & 96.47 & 96.34\\
      \hline
    \end{tabular}
\end{table}

The hyper-parameter $\lambda$ serves as a balancing factor between the length prediction loss $L_{len}$ and the recognition loss $L_{rec}$. We conduct an ablation study on $\lambda$ and present the results in Table~\ref{tab:permutation}.
\end{document}